\pdfoutput=1
\newif\ifdraft\drafttrue
\documentclass[11pt]{article}

\usepackage{acl}

\usepackage{times}
\usepackage{latexsym}
\usepackage{amsmath}
\usepackage{amssymb}
\usepackage{bm}

\usepackage{float}
\restylefloat{table}
\usepackage{times}
\usepackage{latexsym}
\usepackage{amsmath}
\usepackage{amssymb}

\usepackage{float}
\restylefloat{table}
\usepackage{times}
\usepackage{latexsym}
\usepackage{amssymb}
\usepackage{graphicx}
\usepackage{paralist}
\usepackage{graphicx}
\usepackage{subcaption}
\usepackage{caption}
\usepackage{multirow}
\usepackage{multicol}
\usepackage{booktabs}
\usepackage{setspace}
\usepackage{times}
\usepackage[T1]{fontenc}
\usepackage{latexsym}

\usepackage{amsmath}
\usepackage{amsthm}
\usepackage{amsfonts}
\usepackage{xspace}
\usepackage{color}
\usepackage{graphicx}
\usepackage{subcaption}
\usepackage{array}
\newcolumntype{L}{>{\centering\arraybackslash}m{3cm}}

\newcommand*\samethanks[1][\value{footnote}]{\footnotemark[#1]}

\newtheorem{definition}{Definition}

\newtheorem{theorem}{Theorem}

\usepackage[T1]{fontenc}
\usepackage[utf8]{inputenc}

\usepackage{microtype}
\usepackage[compact]{titlesec}
\titlespacing{\section}{0pt}{2ex}{1ex}
\titlespacing{\subsection}{0pt}{1ex}{0ex}
\titlespacing{\subsubsection}{0pt}{0.5ex}{0ex}
\title{Measuring Fairness of Text Classifiers via Prediction Sensitivity}

\author{Satyapriya Krishna \thanks{* Work done while working at Amazon}\\
  Harvard University \\
  \texttt{skrishna@g.harvard.edu} \\ \newline \\
  \textbf{Jwala Dhamala} \\
  Amazon Alexa AI \\
  \texttt{jddhamal@amazon.com} \\\And
  Rahul Gupta \\
  Amazon Alexa AI \\
  \texttt{gupra@amazon.com} \\\newline
  \\
  \textbf{Yada Pruksachatkun \samethanks}\\
  Amazon Alexa AI \\
  \texttt{\small yada.pruksachatkun@gmail.com} \\\And
  Apurv Verma \\
  Amazon Alexa AI \\
  \texttt{verapurv@amazon.com} \\\newline
  \\
  \textbf{Kai-Wei Chang} \\
  UCLA, Amazon Alexa \\
  \texttt{kwchang@cs.ucla.edu}}

\begin{document}
\maketitle
\begin{abstract}
With the rapid growth in language processing applications, fairness has emerged as an important consideration in data-driven solutions. 
Although various fairness definitions have been explored in the recent literature, there is lack of consensus on which metrics most accurately reflect the fairness of a system. 
In this work, we propose a new formulation -- \textsc{accumulated prediction sensitivity}, which measures fairness in machine learning models based on the model's prediction sensitivity to perturbations in input features. 
The metric attempts to quantify the extent to which a single prediction depends on a protected attribute, where the protected attribute encodes the membership status of an individual in a protected group. 
We show that the metric can be theoretically linked with a specific notion of group fairness (statistical parity) and individual fairness. It also correlates well with humans' perception of fairness.
We conduct experiments on two text classification datasets -- \textsc{Jigsaw Toxicity}, and \textsc{Bias in Bios}, and evaluate the correlations between metrics and manual annotations on whether the model produced a fair outcome. We observe that the proposed fairness metric based on prediction sensitivity is statistically significantly more correlated with human annotation than the existing counterfactual fairness metric.

\end{abstract}

\section{Introduction}

Ongoing research is increasingly emphasizing the development of methods which detect and mitigate unfair social bias present in machine learning-based language processing models. These methods come under the umbrella of algorithmic fairness which has been quantitatively expressed with numerous definitions \cite{mehrabi2019survey,jocobs2021measurement}. These fairness definitions are broadly categorized into two types, i.e, individual fairness and group fairness. Individual fairness (e.g., counter-factual fairness \cite{kusner2017counterfactual}) is aimed at evaluating whether a model gives similar predictions for individuals with similar personal attributes (e.g., age or race). 
On the other hand, group fairness (e.g., 
statistical parity \cite{dwork2012fairness}) evaluates fairness across cohorts with same protected attributes instead of individuals \cite{mehrabi2019survey}. 
%Some of the common definitions of group and individual fairness are statistical parity \cite{dwork2012fairness} (group fairness) and counter-factual fairness \cite{kusner2017counterfactual} (individual fairness). 
Although these two broad categories of fairness define valid notions of fairness, human understanding of fairness is also used to measure fairness in machine learning models \cite {dhamala2021bold}. Existing studies often consider only one or two these verticals of measuring fairness. 

%To the best of our knowledge, we are not aware of if existing metrics have been associated with these three verticals of measuring fairness. 

In our work, we propose a formulation based on models sensitivity to input features -- the \textit{accumulated prediction sensitivity}, to measure fairness of model predictions.
%We establish its relationship with specific forms of individual fairness, group fairness and human understanding of fairness. 
We establish its theoretical relationship with statistical parity (group fairness) and individual fairness~\cite{dwork2012fairness} metrics. We then demonstrate the correlation between the proposed metric and human perception of fairness using empirical experiments. 
%We establish its relationship with (i) statistical parity based definition of group fairness using theoretical analysis (ii) individual fairness \cite{dwork2012fairness} using theoretical analysis and (iii) human understanding of fairness using empirical experiments.  

Researchers have proposed metrics to quantify fairness based on a model's sensitivity to input features.
Specifically, \citet{maughan2020towards, ngong2020towards} propose a \emph{prediction sensitivity} metric that attempts to quantify the extent to which a single prediction depends on a protected attribute.
The protected attribute encodes the membership status of an individual in a protected group. 
Prediction sensitivity can be seen as a form of feature attribution, but specialized to the protected attribute.
%The metric also has a counterfactual interpretation as it measures whether or not the model would have made the same prediction, if the individual's protected status were changed. 
In our work, we extend their concept of prediction sensitivity to propose \textit{accumulated prediction sensitivity}.
Akin to the metric proposed by \cite{maughan2020towards, ngong2020towards}, our metric also relies on model output's sensitivity to changes in input features.
Our metric generalizes their notion of sensitivity, where the model sensitivity to various input features can be weighted non-uniformly.
We show that the formulation follows certain properties for the chosen definitions of group and individual fairness and also present several methodologies to select weights assigned to sensitivity of model's output to input features. 
For each selection, we present the correlation between the \textit{accumulated prediction sensitivity} and human assessment of the model-output fairness.

We define our metric in Section ~\ref{section:pred_sense} and present bounds on it (under settings when a classifier follows the selected group fairness or individual fairness constraints) in Sections ~\ref{section:group_fair} and ~\ref{section:indv_fair}, respectively. 
Next, given that the human perception of fairness is not theoretically defined, we present an empirical study on two text classification tasks in Section ~\ref{section:human_fair}.
We request a group of annotators to annotate whether they think that model output is biased against a specific gender and observe that the proposed metric correlates positively with more biased outcomes. 
We then observe correlations between our metric and the stated human understanding of fairness.
We find that not only the proposed accumulated prediction sensitivity metric correlates positively with human perception of bias, but also beats an existing baseline based on counterfactual fairness.

\section{Related Work}
Multiple efforts have looked into defining, measuring, and mitigating biases in NLP models ~\cite{sun2019mitigating,Mehrabi2019ASO,sheng2021defense}. \citet{dwork2012fairness} and \citet{kusner2017counterfactual} focus on individual fairness and propose novel classification approaches to ensure that a classification decision is fair towards an individual. Another set of works focus on group fairness. \citet{corbett2017algorithmic} present fair classification to ensure population from different race groups receive similar treatment. \citet{hardt2016equality} focus on shifting the cost of incorrect classification from disadvantaged groups. \citet{zhao2020logan} measure group fairness in local regions.  
Finally, \citet{kearns2019empirical} combine the best properties of the group and individual notions of fairness.
%on four classification tasks.

Multiple recent works also focus on developing new dataset and associated metrics to capture various types of biases. For example, \citet{dhamala2021bold} and \citet{nangia2020crows} propose dataset and metrics to measure social biases and stereotypes in language model generations, \citet{bolukbasi2016man,caliskan2017semantics,manzini2019black} define metrics to access gender and race biases in word vector representations, and \citet{wang2019balanced} define metrics to quantify and mitigate biases in visual recognition task. \citet{ethayarajh2020is} propose Bernstein bounds to represent uncertainty about the bias. Majority of these bias metrics are automatically computed, for example, using a regard classifier~\cite{sheng2019woman}, sentiment classifier~\cite{dhamala2021bold}, toxicity classifier~\cite{dixon2018measuring} or true positive rate difference between privileged and underprivileged groups~\cite{de2019bias}. A few works additionally validate the alignment of these automatically computed bias metrics with human understanding of biases by collecting annotations of biases on a subset of test data from crowd-workers~\cite{sheng2019woman, dhamala2021bold}. \citet{blodgett2021stereotyping,blodgett2020language} discuss the limitations of several of these bias datasets and measurements. 

However, the majority of existing bias metrics are specific to the model type and the application domain used, they may not be tested for correlation with human judgement of biases, and their relationship to existing definitions of fairness has not been explored. 
{Additionally, metrics such as true positive or error difference between groups requires ground truth labels, thereby making their computation in real-time systems difficult.
\citet{speicher2018unified} have attempted to present unified approach to measuring group and individual fairness via inequality indices, however we note that such metrics are non-trivial to extend to unstructured data such as text.
For example, gender information in a text may be subtle (e.g. mention of softball) and it is unclear whether presence of this word should be considered to impact the genderness of the text.}
\emph{Accumulated prediction sensitivity} metric, presented in this paper, attempts to address all the above limitations of existing bias metrics.
We acknowledge that the proposed metric is yet to be associated with other notions of fairness (e.g. preference based notion of fairness \cite{zafar2017parity}).

% [TODO : I think we add paper on demographic parity, individual fairness , and manual assessment papers such as BOLD, Social Bias Frames. @jwala : please feel free to pick papers form https://arxiv.org/abs/1908.09635 . I am not aware of any paper that follows the same format as us with metric described with respect to all the three corners of fairness (GF, IF, Human labels) ]

\section{Accumulated Prediction Sensitivity}
\label{section:pred_sense}
%\emph{Prediction sensitivity}~\cite{maughan2020towards, ngong2020towards} is a metric that attempts to quantify the \emph{extent to which a single prediction depends on a protected attribute}, where the \emph{protected attribute} encodes the membership status of an individual in a protected class. Prediction can be seen as a form of feature attribution, but specialized to the protected attribute; prediction sensitivity also has a counterfactual interpretation: it measures whether or not the model would have made the same prediction, if the individual's protected status were changed.

Below, we define \emph{accumulated prediction sensitivity}, a metric that captures the sensitivity of a model to protected attributes. 
%We then discuss its relations to group fairness and individual fairness in Sec. \ref{section:group_fair} and Sec. \ref{section:indv_fair}, respectively. 

\begin{definition}[Accumulated Prediction sensitivity]
Let $\bm x \in \bm X$ be a feature vector drawn from the input space $\bm X$. Let
$\bm w, \bm v$ be stochastic vectors whose entries are non-negative values  that sum to one.
Given $\bm x$, let $\bm f$ be a $K$-class classifier, such that $\bm{f}(\bm x) = [f_1(\bm x), .., f_k(\bm x), .., f_K(\bm x)]$ denotes the $K$-dimensional probability output generated by the classifier. We define accumulated prediction sensitivity $P$ as:
\begin{equation}
\label{eq:P}
P = \bm w^T \bm J \bm v; \;\; \text{where} \;\; \bm J(k,i) = \left|\frac{\partial f_k(\bm x)}{\partial x_i}\right|.
\end{equation}
\end{definition}

$\bm J$ is a matrix\footnote{Note that we use the following notation scheme in this paper -- bold capital letters for matrices, bold small letters for vectors and un-bolded letters for scalars.} such that the $(k,i)^{\text{th}}$ entry is $\left|\frac{\partial f_k(\bm x)}{\partial x_i}\right|$, where $x_i$ is the $i^\text{th}$ entry in $\bm x$.  
The product $\bm w^T \bm J$ sums the absolute derivatives $|\frac{\partial f_k(\bm x)}{\partial x_i}|$ across $f_k, k = 1,..,K$ and returns a vector of summed derivatives with respect to each $x_i \in \bm x$. 
The product of $\bm v$ with $\bm w^T \bm J$ further averages the derivatives across all the features $x_i \in \bm x$ to yield the scalar $P$. 

The value $\frac{\partial f_k(\bm x)}{\partial x_i}$ captures the expected change in model output for the $k^\text{th}$ class given a perturbation in $x_i$.
If $x_i$ is a protected feature, arguably a smaller value of $\frac{\partial f_k(\bm x)}{\partial x_i}$ implies a fairer model; as then the model's outcome does not change sharply with changes in $x_i$.
To capture the sensitivity of the model with respect to the protected features, one also needs to choose $\bm v$ judiciously. 
For example, given the explicit set of protected features in $\bm x$, one can select $\bm v$ such that only entries corresponding to those features are assigned a non-zero value, while the rest are set to zero.
Given this heuristic, we expect the value $P$ to be smaller for fairer models.
%We discuss our methodology to select $\bm w$ and $\bm v$ in the experiments section. 
In the following sections, we connect the accumulated prediction sensitivity to two known notions of fairness and human perception of fairness.

%\paragraph{Formal Definition.}
%The original formalization of prediction sensitivity is as follows.
%A model $\mathcal{F}$ makes a prediction $\hat{y}$ for an example $x$ based on model weights $\theta$ as $\hat{y} = \mathcal{F}(\theta, x)$. Let $Z$ be the set of tokens in the example that refer to a protected attribute. 
%The prediction sensitivity for $\mathcal{F}$ on $x$ with respect to the protected attribute $z \in Z$ is defined to be the following the partial derivative:
%
%\[ \Big\lvert \frac{\partial}{\partial z} \mathcal{F}(\theta, x) \Big\rvert \]
%
%This is a scalar value that can be efficiently computed in most deep learning frameworks. A high value for prediction sensitivity indicates that bias was a likely factor in the individual prediction it is associated with.

% \paragraph{Prediction Sensitivity as a Fairness Metric.}
% \textbf{TODO: do we want this section? What do we say?}

\section{Relation to Group Fairness: Statistical Parity}
\label{section:group_fair}
%We connect our definition of prediction sensitivity to two existing definitions of fairness. 
Given a set of protected features (e.g. gender), a model satisfies statistical parity if model outcome is independent of the protected features (we note that identifying protected features may not always be feasible in the real world).
We represent the feature vector $\bm x = [\bm x_p, \bm x_l]$, where $\bm x_p$ is the set of protected features and $\bm x_l$ is the remainder. 
Accordingly, we choose $\bm v$ to be a vector such that the entries that sum $|\frac{\partial {f_k}(x_p)}{\partial  x_i}| \forall x_p \in \bm x_p$ in $\bm J$ are non-zero; and zero otherwise. 
This choice is intuitive as then we sum the gradients in $\bm J$ that correspond to protected features and measure model's sensitivity to them.
The predictor ${\bm f}(\bm x)$ will satisfy statistical parity if $\bm{f}(\bm x_p, \bm x_l) = \bm{f}(\bm x_p^\prime, \bm x_l) \forall \bm x_p \neq \bm x_p^\prime$. Given this, we state the following theorem.

\begin{theorem}
Given a vector $\bm v$ with non-zero entries corresponding to $\bm x_p$ and zero entries for $\bm x_l$, if the predictor $\bm{f}(\bm x)$ satisfies statistical parity with respect to $\bm x_p$, accumulated prediction sensitivity will be zero. 
\end{theorem}

{\bf Proof}: If $\bm{f}(\bm x)$ satisfies statistical parity with respect to $\bm x_p$, the values $\frac{\partial {f_k}(\bm x)}{\partial  x_p}  \forall x_p \in \bm x_p$ will be all zeros. 
This is due to the fact that the function $f_k(\bm x)$ can not be defined based on entries $x_p \in \bm x_p$ for it to be independent of them.
Therefore, for every multiplication in the product $\bm J \bm v$, either the entry $\frac{\partial {f_k}(\bm x)}{\partial  x_p}$ will be 0 or the entry in $\bm v$ corresponding to $\bm x_l$ will be 0. Hence, $P$ will be 0.

Appendix A presents empirical results in computing $P$ on a synthetic dataset. 
We construct a dataset where a feature (hair length) correlates with a protected attribute (gender).  
We show that if the modeler unintentionally uses the correlated feature while attempting to build a classifier with statistical parity, our metric can be used for evaluation.

\section{Relation to Individual Fairness}
\label{section:indv_fair}
\citet{dwork2012fairness} state the notion of individual based fairness as: "{\it We interpret the goal of mapping similar people similarly to mean that the distributions assigned to similar people are similar}".
They propose adding a Lipschitz property constraint during the classifier optimization. 
Given a loss function $\mathcal{L}$ defined to optimize the parameters $\bm \theta$ of the classifier $\bm{f}(\bm x)$, a distance function $ d(\bm x, \bm x^\prime)$ that computes distance between data-points $\bm x, \bm x^\prime$, another distance function $\mathcal D(\bm f(\bm x)), \bm f(\bm x^\prime))$ that computes distance between classifier predictions on $\bm x, \bm x^\prime$ and a constant $L$, \citet{dwork2012fairness} propose the following constrained optimization. 
\begin{equation}\label{eq:const_opt}
\begin{aligned}
    &\min_{\bm \theta} \mathcal{L}; \quad \text{such that}  \\
    & \mathcal D(\bm f(\bm x)), \bm f(\bm x^\prime)) < L d(\bm x, \bm x^\prime); \forall \bm x, \bm x^\prime \in \bm X.
\end{aligned}
\end{equation}

It is natural to choose an Lp norm \cite{bourbaki1987topological} for $d$ and $\mathcal D$. 
%Note that one may re-define the constraint in Eq.~\eqref{eq:const_opt} for ``mapping similar people similarly'' as $D(\mathcal{F}(\bm x)), \mathcal{F}(\bm x^\prime)) < L d(\bm x_p, \bm x_p^\prime)$, such that the change in model prediction is bounded by the changes in protected features one. 
%We derive a bound of $P$ with the formulation in Eq.~\eqref{eq:const_opt} with the note that it will also apply if the constraint was reformulated to apply only to protected features. 
For a classifier $\bm f$ that is trained with the above constrained optimization and the choice of distance metrics $\mathcal D, d$ is an Lp norm, we state the following. 
\begin{theorem}
If the predictor $\bm{f}(\bm x)$ is trained with the constrained optimization stated in Eq.~\eqref{eq:const_opt}, the accumulated prediction sensitivity will be upper bounded by $L$.
\end{theorem}

{\bf Proof}: We restate the constraint in Eq.~\eqref{eq:const_opt} as (Note that the inequality sign does not change as distance metrics $\mathcal D, d$ are required to be positive for $\bm x \neq \bm x^\prime$)
\begin{equation}\label{eq:lip_re}
    \forall \bm x \neq \bm x^\prime, \quad L > \frac {\mathcal D(\bm{f}(\bm x), \bm{f}(\bm x^\prime))} { d (\bm x, \bm x^\prime)}. 
\end{equation}

Given the inequality holds for any pair of $\bm x, \bm x^\prime$, it must also hold for an $\bm x^\prime$ of the following choice.
%\begin{equation*}
 $   \bm x^\prime = \bm x + [0, 0, \Delta x_i, 0, 0],$
%\end{equation*}
where $\Delta x_i$ is a scalar perturbation in the $i^\text{th}$ entry in $\bm x$. For a chosen Lp norm, Eq~\eqref{eq:lip_re} becomes
\begin{eqnarray}
    L &>& \frac {[\sum_{k=1}^K |f_k(\bm x) - f_k(\bm x^\prime)|^p]^{\frac{1}{p}} } {|\Delta x_i|}  \nonumber \\
   &>& \frac {[|f_k(\bm x) - f_k(\bm x^\prime)|^p]^{\frac{1}{p}}} {|\Delta x_i|}. \label{eq:lip_re2}
\end{eqnarray}

Since each entry $|f_k(\bm x) - f_k(\bm x^\prime)|^p, k=1,..K$ is expected to be non-zero and zeroing out all such entries (but one) will yield a lower value than the summation $\sum_{k=1}^K |f_k(\bm x) - f_k(\bm x^\prime)|^p$. 
We can re-write Eq.~\eqref{eq:lip_re2} as:
\begin{equation*}\label{eq:derivative}
   \frac {|f_k(\bm x) - f_k(\bm x + [0, 0, \Delta x_i, 0, 0])|} {|\Delta x_i|}.
\end{equation*}

We can further chose $\Delta x_i$ such that it is small perturbation, leading to the following. 
\begin{equation*}\label{eq:derivative}
\begin{aligned}
   L >  & \lim_{\Delta x_i \rightarrow 0} \frac {|f_k(\bm x) - f_k(\bm x + [0, 0, \Delta x_i, 0, 0])|} {|\Delta x_i|} \\
   = & \Big |\frac{\partial f_k(\bm x)}{ \partial x_i} \Big |.
\end{aligned}
\end{equation*}

Therefore, each entry in $\bm J$ is upper bounded by $L$.
As vectors $\bm v, \bm w$ are stochastic and they compute weighted averages of bounded entries in $\bm J$, $P$ (defined in Eq. \eqref{eq:P}) must be less than or equal to $L$. 

We also note that as $L$ becomes larger, the constraint in the Eq.~\eqref{eq:const_opt} becomes looser.
Therefore, a higher value of $L$ during optimization is expected to loosen the fairness constraint as well as the bound on fairness sensitivity.
This aligns with our intuition of lower values of $P$ for fairer models. 
We compute value of $L$ on a synthetically generated classification data, optimized with the individual fairness constraint in equation \ref{eq:const_opt}.
The results are presented in Appendix B.

\section{Correlations with Human Perception of Fairness}
\label{section:human_fair}
While the conditional statistical parity and individual fairness establish theoretical constraints on the model behaviour (e.g. independence from protected features and similarity in prediction outcomes for similar data-points), humans may carry a different notion of fairness for model outcomes on individual data-points. 
This notion may be based on their understanding of cultural norms, which in turn effect their decisions in identifying which model outputs could be considered biased. 
In this section, we present experiments that correlate accumulated prediction sensitivity with human perception of fairness.

\subsection{Human Perception of Fairness}
Given a data-point $\bm x$ and model prediction $\bm f(\bm x)$, we assign one of the $K$ classes to the data-point. 
In order to evaluate the human perception of fairness on the data-point, we request a group of annotators to evaluate the model prediction (taken as the arg-max of the model output) and assess whether they believe the output is biased. For instance, given the social/cultural norms, a profession classifier assigning a data-point ``she worked in a hospital'' to nurse instead of doctor can be perceived as biased.
%\section{Experiments}
% \textbf{\textcolor{red}{Add experiment details including details about dataset, metrics, and MTurk experiment design : @satya}}
% \textbf{\textcolor{red}{Add jigsaw dataset experiment details , how was the samples chosen for annotation, and how were they filtered.  : @satya}, Experiment TODO : get correlation numbers for gender filtered samples and overall (already added) Add screenshot of the mturk experiment setup} 
To correlate the accumulated prediction sensitivity $P$ with the human understanding of fairness, we conduct experiments on two text classification datasets. 
We describe the datasets below, followed by our choices for $\bm w$ and $\bm v$.
%For each dataset, we trained two models, i.e, (1) Task Classifier(TC) which is the classifier trained to perform the underlying task, and (2) Gender Classifier which plays the role of PSM defined in section ~\ref{section:psm}. These two classifiers are then used to compute the WPS using equation ~\ref{eq:wps_val}. The resulting WPS score is then used to compute correlation with human annotations. The  details about datasets, experimental fairness metrics, and evaluations are given in the subsections below.  

\subsection{Datasets}
%[TODO : Add details about hyper-params ]

We experiment with our proposed metric on two classification tasks, i.e, occupation classification on \emph{Bias in Bios} dataset ~\citep{DeArteaga2019BiasIB}\footnote{The data is available at https://github.com/microsoft/biosbias}  and toxicity classification with \emph{Jigsaw Toxicity} dataset\footnote{The data is available at https://www.kaggle.com/c/jigsaw-unintended-bias-in-toxicity-classification}. 
{We focus on these two datasets as they have been investigated in several previous studies \cite{pruksachatkun2021does} and have been reported to carry significant presence of  bias.}
%~\citep{DeArteaga2019BiasIB}}. 
\textsc{Bias in bios} data ~\citep{DeArteaga2019BiasIB} is purposed to train occupation classifier which predicts occupation given the biography of an individual. For this data, the task classifier is an occupation classification model which is composed of a standard LSTM-based encoder combined with the output layer of 28 nodes, i.e, number of occupation classes. 
%We use the same model architecture for PSM trained with gender annotations in the dataset. TC trained with the dataset achieved an accuracy of 81.37\% and PSM achieved an accuracy of 98.79\%.
\textsc{Jigsaw Toxicity} dataset is commonly used to train toxic classifier which is tasked to predict if an input sentence is toxic or not. This dataset has input sentences as the comments from Wikipedia's talk page edits labeled with the degree of toxicity. In this dataset, the task classifier is a binary classifier trained to predict whether a comment is toxic or not. We labeled the samples with >0.5 toxicity score as \textit{toxic} and others as \textit{non-toxic} to train the task classifier. The task classifier trained with \emph{Jigsaw Toxicity} dataset achieved an AUC of 0.957. Table \ref{table:dataset_stats} in appendix summarizes the train/test/valid split for the 2 datasets.

\begin{table*}[]
\centering
\small
\begin{tabular}{p{7cm}p{1cm}p{1cm}|p{1cm}p{1cm}p{1cm}p{1cm}p{0.9cm}}
%\toprule
%\multirow{1}{*}{Individual Fairness Metrics} & \multicolumn{3}{l}{Point-biserial correlation ($\downarrow$)} & \multicolumn{4}{l}{ Mutual Information ($\downarrow$)} \\
%\midrule
% & BiB & JT(G) & JT(R) & BiB & JT(G) & JT(R) & JT(GR)  \\ \cline{2-8}
%P1 (uniform $\bm w, \bm v$) & 0.206 & 0.117 & 0.064 & 0.013 & 0.007 & 0.009 & 0.011 \\ 
%CF \cite{garg2019counterfactual} & 0.326 & 0.214 & 0.098 & 0.025 & 0.022 & 0.015 & 0.029  \\ 
%P4 ($\bm v$ set using gendered words) & 0.34 & 0.227 & 0.102 & 0.037 & 0.054 & 0.007 & 0.031 \\ 
%P5 ($\bm v$ set using gendered words and embedding vectors) & \textbf{0.363} & 0.295 & 0.108 & \textbf{0.098} & 0.061 & 0.027 & 0.042 \\ 
%P2 ($\bm v$ set using PSM)& \textbf{0.397} & 0.358 & 0.113 & \textbf{0.102} & \textbf{0.097} & 0.021 & 0.077 \\ 
%P3  ($\bm v$ set using PSM and embedding vectors) & \textbf{0.441} & \textbf{0.374}& 0.101 & \textbf{0.105} & \textbf{0.101} & 0.038 &0.084\\ 
%\bottomrule

\toprule
\multirow{1}{*}{Individual Fairness Metrics} & \multicolumn{2}{c|}{Bias in Bios} & \multicolumn{2}{c}{Jigsaw Toxicity} \\
\midrule
 & Corr. & MI & Corr. & MI  \\ \cline{2-5}
P1 (uniform $\bm w, \bm v$) & 0.206 & 0.013 & 0.117 & 0.007 \\ 
CF \cite{garg2019counterfactual} & 0.326 & 0.025 & 0.214 & 0.022  \\ 
P4 ($\bm v$ set using gendered words) & 0.34 & 0.037 & 0.227 & 0.054 \\ 
P5 ($\bm v$ set using gendered words and embedding vectors) & \textbf{0.363} & \textbf{0.098} & 0.295 & 0.061  \\ 
P2 ($\bm v$ set using PSM)& \textbf{0.397} & \textbf{0.102} & 0.358 &  \textbf{0.097}  \\ 
P3  ($\bm v$ set using PSM and embedding vectors) & \textbf{0.441} & \textbf{0.105}&  \textbf{0.374} &  \textbf{0.101} \\ 

%Numbers with Bios in Bias PSM (Male test set accuracy : 98.84, Female test set accuracy : 98.17)
%Numbers with Jigsaw Toxicity PSM (Male test set accuracy : 95.52, Female test set accuracy : 96.22)

\bottomrule

\end{tabular}
\caption{\label{citation-guide} Point bi-serial correlations (Corr.) and Mutual Information (MI) between different individual fairness metrics with human annotations on Bios in Bias and Jigsaw toxicity datasets. Bold numbers are the correlations where we see statistically significant increase over CF baseline. The metric variants are sorted based on the correlation values. We use the bootstrap method to compute statistical significance \cite{koehn2004statistical} at p-value<0.05.}
\label{tab:table1}
\end{table*}

\subsection{Selecting the vectors $\bm w$}
The vector $\bm w$ sums up the absolute partial derivatives of $f_k(\bm x)$ with respect to a given feature $x_i, \forall k=1,..,K$.
In our setup, we consider input features to be the word embeddings and the matrix $\bm J$ is computed over the same. Given a $D$-dimensional word embedding, $K$ classes and $N$ words in $\bm x$, $\bm J$ will be a matrix of size $(K)\times(DN)$. 
In all our experiments, we choose $\bm w$ to be a uniform vector with entries $1/K$.
Such a choice assigns equal weight to the partial derivatives computed over each class.
One may chose to put a higher weight on derivatives computed over a specific class, if there is a reason to believe that the accumulated prediction sensitivity should be informed more with respect to that class.
For instance, for a classifier that stratifies medical images into various diseases \cite{agrawal2019evaluating}, disparity in model performance with respect to malicious diseases can be considered more costly. 
Therefore, derivatives for classes that represent more malicious disease can be weighted higher. 

%\subsubsection{Uniformly weighted $\bm w$}
%In this setup, the values of $\bm w$ are set to $1/(DN)$, therefore assigning an equal weightage to all partial derivatives computed w.r.t. a given feature $x_i$.

%\subsubsection{Using the embedding vectors as $\bm w$}
%\cite{han2020explaining} proposed a methodology that compute saliency maps over  features that multiplies the partial derivatives with respect to embeddings with the embedding values themselves (equivalently $\bm J$ with $\bm w$ in our work). 
%Inspired by their work, we also set the values in $\bm w$ equal to the embedding vectors. 

\subsection{Selecting the vectors $\bm v$}
\label{section:psm}

Through the vector $\bm v$, we aim to select words in $\bm x$ that carry gendered information. We use two formulations for the the vector $\bm v$ as discussed below.

\subsubsection{Using a list of gendered words}\label{sec:gendered_words}
In this setup, we use the set of gendered words from \cite{bolukbasi2016man} and assign entries in $\bm v$ corresponding to those words as $1/(N_g \times D)$, where $N_g$ is the count of gendered words in the data-point. 

\subsubsection{Using a Protected Status Model (PSM)}
While prior work has used word matching to a pre-defined corpus of tokens describing various demographic cohorts \cite{bolukbasi2016man}, these corpus do not contain words that stereotypically are associated with a particular cohort but may not be explicitly tied to that cohort. For example, the word ``volleyball'' is associated with females in the analysis presented by \cite{dinan2020multi}.

To capture this nuance, we propose using another classifier (that acts on the same dataset as used to train the original classifier, for which we aim to compute $P$) and using it to identify tokens containing information about the protected attribute (e.g. gender). We discuss the model training below. 

\paragraph{Protected Status Model:}
To extend accumulated prediction sensitivity to settings with no explicit protected attribute, we train a \emph{protected status model} $\bm g$. 
%whose goal is to predict the protected status of individuals. 
%Formally, training this model results in weights $\theta_2$ such that $\hat{z} = \mathcal{A}(\theta_2, x)$ and $\hat{z}$ approximates the true protected attribute $z$.
Given the data-point $\bm x$, goal of the PSM model $\bm g(\bm x)$ is to predict the protected attributes. 
%We train PSM on a dataset that assigns the protected attribute to features (in the same feature space as the original model $\bm f$). 
Given a trained $\bm g(\bm x)$, we then compute another matrix $\bm J_g$, where the $(j,i)^\text{th}$ entry is $|\frac{\partial \bm g_m(\bm x)}{\bm x_i}|$ ($\bm g_m$ is the probability outcomes corresponding to the $m^\text{th}$ protected attribute class; e.g. male in a gender classifier). 
We then define an entry $v_i \in \bm v$ as $\sum_j \bm J_g(m,i)$ (the vector $\bm v$ is normalized to be stochastic).
Intuitively, the sum $\sum_j \bm J_g(m,i)$ captures the model output sensitivity with respect to the input features $x_i$ and is expected to higher if $x_i$ carries more gendered information. 

In our experiments, we train separate PSM models for gender sensitivity computation on Bias-in-bios and Jigsaw data-sets, as each data-point in these data-sets is additionally labeled with a binary gender class (male/female)\footnote{We note that this is a limitation of this work as gender can be non-binary.}. 
%Similarly, we train a PSM model for race sensitivity computation on the Jigsaw dataset. 
Gender PSMs predicts the associated gender given the datapoint $\bm x$.
Training PSM on the same datasets used to train the task classifier $\bm f$ helps capture the gender stereotypes present in the respective datasets.
For instance, in a given dataset, if the word ``volleyball'' appears more often in the data-points that correspond to the female gender, the gender classifier's sensitivity to this word is expected to be high as the classifier may pay higher emphasis to this word for gender classification.
We use the same model architecture as the task classifiers for PSM. PSM for gender classification achieve an accuracy of 98.79\% (Male Acc:98.84\% Female Acc:98.17\%) and 95.39\% (Male Acc:95.92\% Female Acc:96.22\%) for {\it Bias in bios} and {\it Jigsaw Toxicity} datasets, respectively. 
%PSM model for race classification in the {\it Jigsaw Toxicity} dataset achieves an accuracy of x\%. 
These accuracies are computed over the same train/test split as the task classifier.

\subsubsection{Using Word Embedding Vectors}
In addition to using the list of gendered words and PSM, we also test with a setting where we multiply the word embedding vectors to the proposed formulations of $\bm v$.
We stack the word embedding vectors for each word $x_i \in \bm x$ to obtain a vector of embeddings $\bm e_i$. 
We perform an element-wise multiplication of the embedding vectors $\bm e_i$ with the vector with entries $1/(N_g \times D)$ for gendered words or $\sum_j \bm J_g(j,i)$ obtained using PSM.
This choice is motivated based upon the findings in \cite{han2020explaining}.
They leverage the magnitude of embedding vectors in determining saliency of the input words for the classification task at hand.
Their proposed methodology computes saliency maps over the features $x_i \in \bm x$ by multiplying embedding vectors with partial derivatives of the class probabilities with respect to embedding vectors themselves.

\subsection{Fairness Metrics}

We experiment with six fairness metrics. 
%and use correlations with over 900 human annotations for Bias in Bios dataset, as well as Jigsaw toxicity to evaluate their alignment with the human understanding of fairness. 
Out of the six, one metric is a baseline based on counter-factual fairness and the rest are variants of the accumulated prediction sensitivity $P$. %They are listed below.
%\begin{itemize}

  \noindent \textbf{Counter-factual Fairness (CF)} : We use the counter-factual fairness definition mentioned in \citet{garg2019counterfactual} and compute the metric as the difference in model predictions between the original sample $\bm f(\bm x)$ and its corresponding counter-factual gendered sample $\bm f(\hat{\bm x})$. We take the L1 norm of the vector $\bm f(\bm x) - \bm f(\hat{\bm x})$. For example, we take the difference in predictions between the sample "She practices dentistry" and "He practices dentistry", which is the corresponding counter-factual sample. We use the definitional gender token substitutions from \citet{bolukbasi2016man} to create counter-factual samples. 
  
   \noindent \textbf{P1: Uniformly weighted prediction sensitivity} : In this setting, the values of $\bm w$ and $\bm v$ are set to uniform values $\frac{1}{K}$ and $\frac{1}{DN}$, respectively. 
  This is a weak baseline as the choice of $\bm v$ does not provide any information regarding the gender-ness of the input words. 
   
  \noindent \textbf{P2: Weighted Prediction Sensitivity based on PSM} : In this setting, $\bm w$ is chosen to be a uniform vector, while $\bm v$ is chosen based on the PSM model. 
  
  \noindent \textbf{P3: Weighted Prediction sensitivity + Embedding weights} : In this setting, $\bm v$ is chosen based on the PSM model (akin to the metric in P2) which is further multiplied element-wise with the word embedding vectors. 
  
  \noindent \textbf{P4: Hard gender weights based Prediction sensitivity} : In this metric, we use the list of gendered words described in section~\ref{sec:gendered_words} to determine $\bm v$. The value of entries in $\bm v$ is set to $\frac{1}{DN_g}$.
  
  \noindent \textbf{P5: Hard gender weights based prediction sensitivity + Embeddings}: This setting is same as above, except entries in $\bm v$ are further multiplied element-wise with the word embedding vectors. 
  
%\end{itemize}

\subsection{Evaluation}

\begin{table*}[t]
\centering 
\small
\begin{tabular}{c c}
\toprule
\multicolumn{2}{c}{\textbf{Example from the Bias in Bios dataset}}\\
TC &  \includegraphics[width=0.9\textwidth]{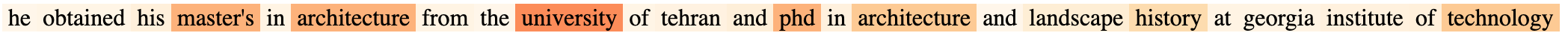} \\
PSM & \includegraphics[width=0.9\textwidth]{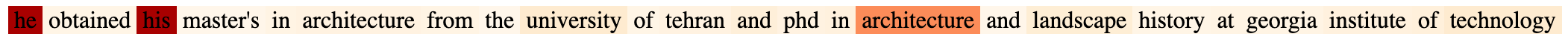} \\
\midrule
\multicolumn{2}{c}{\textbf{Example from the Jigsaw Toxicity dataset}}\\
TC & \includegraphics[width=0.9\textwidth]{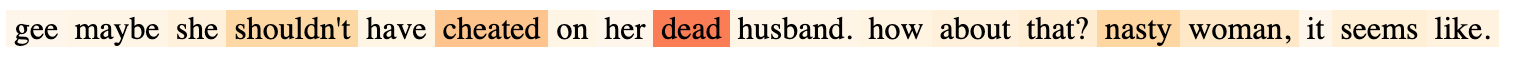} \\
PSM & \includegraphics[width=0.9\textwidth]{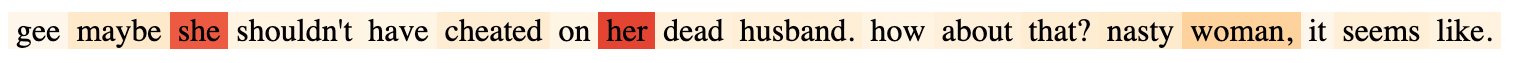} \\
\bottomrule
\end{tabular}
\caption{Heat map for the vectors $\bm w^T J$ (top entry in each row) and $\bm v$ (bottom entry in each row) per input word $x_i$. Darker color indicates higher magnitude. These vectors are multiplied to compute \textsc{accumulated prediction sensitivity}. \textbf{TC}: Task Classifier, \textbf{PSM}: Protected Status Model.} 
\label{table:saliency}
\end{table*}

To evaluate whether the proposed prediction sensitivity correlates with human perception of fairness, we collect annotations from crowd workers using the Amazon Mechanical Turk platform. Crowd workers are asked to annotate if a model prediction appears to be a biased prediction or not. For \emph{Bias in Bios} dataset, each sample presented to the annotators has the biography and occupation predicted by the model. We collect annotations on a random sample of the test set. For each biography and a predicted occupation, we ask annotators to label if the prediction is indicative of bias or if it is unbiased. Bias refers to a situation where an occupation is incorrectly predicted based on the gender associated with the biography. For instance, if the input biography is ``she studied at Harvard Medical School and practices dentistry.'' and is predicted as nurse, then we call this prediction biased since the biography fits better for a doctor. 
%In the cases of weak bias, an occupation is predicted based on the gender of the biography, however, the biography may not clearly indicate a profession. 
In case of unbiased predictions, the prediction is not expected to be influenced by the gender content in the biography. Table 3 presents  a sample of examples provided to the annotators for the \emph{Bias in bios} dataset.
Each page in the annotation task consisted of ten biography-profession pairs. We collect annotations for each biography-profession pair from at least three annotators and pick the label with majority vote. Similarly for \emph{Jigsaw Toxicity} dataset, each sample presented to the annotators contains the text and associated toxicity predicted by the model. 

% Created a table instead for this figure
% \begin{figure*}[t!]
%     \centering
%     \includegraphics[width=16cm, trim={0 0 0 7cm},clip]{images/Screen Shot 2021-06-14 at 10.14.01 PM.png}
%     \caption{Examples of biased/unbiased outcomes shown to the M-turk annotators}
%     \label{fig:Mturk}
% \end{figure*}

\begin{table}[t]
\label{table:MTurk_biasbios}
    \centering\small
    \begin{tabular}{@{}p{\columnwidth}@{}}
    \toprule
        \textbf{EXAMPLES OF UNBIASED SAMPLES} (The predicted profession is unrelated to gender stereotype about professions)\\\midrule
        \textbf{BIO}: She received a master's degree in computer science from the university of North Carolina at Chapel Hill.\\ \textbf{Predicted Profession}: Computer Scientist \\
        \\
        \textbf{BIO}: He received a master's degree in computer science from the university of North Carolina at Chapel Hill.\\ \textbf{Predicted Profession}: Computer Scientist \\\midrule
        \textbf{EXAMPLES OF BIASED SAMPLES} (Strongly biased predictions are based on associating a specific gender to a specific profession even when there are evidences against it in the biography)\\\midrule
        \textbf{BIO}: Mary has 25 years of experience in data analytics, business intelligence and information governance with fortune 100 companies.\\
        \textbf{Predicted Profession}: Nurse \\
        \\
        \textbf{BIO}: He achieved a masters degree in nursing from the university of north Carolina at chapel hill\\
        \textbf{Predicted Profession}: Computer Scientist\\
    \bottomrule
    \end{tabular}
    \caption{Examples of biased/unbiased outcomes shown to the M-turk annotators}
\end{table}

%Unlike \emph{Bias in Bios}, we only have two labels for human annotation of \emph{Jigsaw Toxicity} samples, i.e, "biased" and "unbiased". In order to reduce noise in human annotations, we only annotate comments with word length > 10. 
We restrict the set of annotators to be master annotators and the location of annotators to be Unites States. Based on the initial pilot studies conducted in the Amazon Mechanical Turk platform, we setup a payment rate to ensure a fair compensation of at least $15\$/hour$ for all annotators that work at an average pace.  We annotated 900 test data-points from each dataset.
We note that these test data-points were misclassified by the classifiers $\bm f$ trained for each dataset. 
While such a sampling may not conform to the true distribution of biased/unbiased model outcomes on the overall test set, we expect to get more biased samples amongst the misclassified samples. 
The distribution between biased and unbiased outputs was about 55:45 for \emph{Bias in Bios} and 50:50 for \emph{Jigsaw Toxicity}. 
For the \emph{Bias in Bios} and \emph{Jigsaw Toxicity} datsets, we obtained a Fliess' kappa of 0.43 and 0.47, respectively, amongst the three annotators.
{This is considered a moderate level of agreement, which we believe is expected for an relatively ambiguous task to identify model outcomes influenced by gender.}
We compute mutual information and bi-serial correlations as the primary measures of association between the human annotations and the {\it accumulated model sensitivity}.

\section{Results}

%\textbf{TODO : Insert table, MTurk results, and qualitative examples : @satya }

% Our results in Table \ref{tab:template} show that weighted prediction sensitivity and its variants are strongly correlated with in fairness metric for three occupations that exhibits most bias \cite{de2019bias}.   

% \subsection{Correlations with Individual Fairness}
Table \ref{tab:table1} lists the bi-serial correlations and mutual information between manual annotations and the different fairness metrics. 
First, we observe that correlations of the baseline with human judgement are mediocre (0.326 and 0.214) compared to the human judgement.
We attribute this to the fact that the metric attempts to quantify a fairly subjective assessment of bias that may have different interpretation (as also pointed out by the moderate level of annotation agreement across annotators).
However, the proposed variants of $P$ have stronger correlations compared to the counter-factual baseline (except the method P1). 
As expected, we see the smallest correlation for P1, since this metric does not account for gender-ness in $\bm v$. 
However, metrics that determine $\bm v$ based on PSM prediction sensitivity and gendered words get higher correlations over P1 and the CF baseline.
Variant of $P$ with $\bm v$ informed using the embedding vectors further lead to improved correlations. 
We also observe weaker statistical significance in the case of \emph{Jigsaw Toxicity} due to a weaker PSM.
We attribute this to the noise present in gender annotations for \emph{Jigsaw Toxicity} dataset. Hence, the performance of PSM in predicting the protected status is crucial for accurately measuring fairness. 

\subsection{Discussion}

In order to further analyse the effect of PSM, we look into heat-maps capturing $\bm w^T \bm J$ and $\bm v$ separately.
As a reminder, the first quantity captures the weighted average of partial derivatives of class probabilites with respect to the input features, while the second quantity computes the weights assigned to sum up the aforementioned averages.
Table \ref{table:saliency} shows while $\bm v$ mostly captures gendered words such as ``she'', ``her'' and ``woman'', it also captures words such as ``social'', ``architecture'' and ``cheated'' to carry more gendered information compared to other words. 
While these words conventionally are not gendered, for the datasets at hand, they seem to provide information whether the input data-point belongs to male/female gender.
We also note that $\bm w^T \bm J$ weighs on occupation specific tokens such as "physician", "executive", etc.

%When these weights are combined with dot-product, it tends to amplify gender affinity of a sample and hence, resulting in improved measurement fairness of a sample. 

%We noticed similar pattern in samples from \emph{Jigsaw Toxicity} where TC assigns higher score to toxic words present in the sample such as "cheated" and "nasty" in example 3 in Table ~\ref{table:saliency}, whereas PSM gives higher weight to gendered tokens, hence, combining these two weights gives a reliable measure of fairness. 

This finding supports our motivations to compute $\bm v$ based on PSM and capturing feature attributions assigned to tokens that are implicitly related to a specific gender (instead of the definitional gender tokens only). 
%This can be seen in second example in Table \ref{table:saliency} where PSM gives higher weight to the token "architecture" for a male related biography since the word "architecture" is used more often in the context of male over females. 
Hence, by incorporating PSM in computing $P$, we can capture bias present in non-trivial gendered tokens. 
%This also justifies the better correlation achieved by WPS over metrics using hard gender weights in ~\ref{tab:table1}. 

%\subsection{Correlations with Group Fairness}

% Table \ref{tab:template} shows the TPR difference scores along with weighted prediction sensitivity for the top three most biased occupations, where we see strong correlations for all the occupations. 

%Our results in Table \ref{tab:template} shows the weighted prediction sensitivity score and TPR difference for the top three most biased occupations observed in \citet{de2019bias}. We notice that the application of debiasing method does bring down the sensitivity scores similar to TPR difference. This trend supports the argument that weighted prediction sensitivity is sensitive to unfair bias present in the model, and hence can measure relative changes in model fairness. This trend was noticed for all the 28 occupations. However, we dont observe direct correlations with TPR difference as we notice lowest prediction sensitivity score for "Model" occupation compared to that of "Nurse", which is against the trend for TPR difference. One of the possible reasons behind this trend could be the mis-alignment in the the definitions followed by TPR difference, which is based on Equality of Opportunity (EOp), whereas our proposed metric is based on sensitivity with respect to protected attributes. 
%\section{[WIP]Related Work}
% Add papers with related work 

\section{Considerations for Accumulated Prediction Sensitivity Metric}
While the results showcase the promise of our metric, we draw the attention of the reader to the following considerations: (1) We observed that the metric quality depends on choice of the hyper-parameters $w$ and $v$. 
In this regard, our metric is not different from other metrics that also depend on a hyper-parameter choice. 
For example, any classifier based metric has a threshold parameter and counterfactual fairness metrics rely on hyper-parameters such as the selected gendered words. (2) Our metric only works for models for which gradients can be computed. Most modern deep learning based models carry this property. (3) Lastly, we note that it is hard to interpret the absolute value of the proposed metric. The metric value should be used for relative comparison of two models which share input feature space and label space. 

In addition, we note two considerations for relying on a PSM classifier. First, training it requires access to gender labels. Second, the PSM model itself could be biased. 
Given that gender labels may not always be available for the dataset used to train model at hand, we study the impact of transferring a PSM model trained on a different dataset on computing our metric.
We also evaluate the effect of bias in PSM model on the overall metric value and present results in the Appendix ~\ref{section:appendix_psm_classifier}. 
We make observations such as the quality of the metric degrades as PSM becomes more biased. 
%These findings necessitate careful consideration in the choice of a PSM model.
Based on these observations, we recommend that if modeler is not able to obtain high performance PSM models, they fall back to using sources such as gendered words for computing the vector $\bm v$.

\section{Conclusion}
Evaluating fairness is a challenging task as it requires selecting a notion of fairness (e.g. group or individual fairness) and then identifying metrics that can capture these notions of fairness while evaluating a classifier. 
Additionally, certain notions of fairness may not be well defined and can change based upon social norms (e.g. ``volleyball'' being closely associated with females); that may seep into the dataset at hand. 
In this work, we define an accumulated prediction sensitivity metric that relies on the partial derivatives of model's class probabilities with respect to input features. 
We establish properties of this metric with respect to the three verticals of fairness metrics: group, individual and human-perception based.
We provide bounds on the metric's value when a predictor is expected to carry statistical parity or is trained with individual fairness. 
We also evaluate this metric with fairness as perceived through human evaluation of model outputs.  
We test variants of the proposed metric against an existing baseline derived from counter-factual fairness and observe better mutual information and correlation.
Specifically, a variant of the metric that relies on a Protected Status Model (that identifies tokens that carry gender information but may not conventionally be considered gendered) yields the best correlation with the human evaluation.

In the future, one can associate the proposed formulation with other categories of group and individual fairness \cite{Mehrabi2019ASO}. 
We also aim to test the metric on other datasets with other protected attributes (e.g. race, nationality). 
Finally, we can compare the metric across these datasets to compare trends across protected groups.

\section{Broader Impact and Ethics Statement}
This work can be used to evaluate bias in models, and thus used to evaluate models serving human consumers. As with all metrics, the metric does not capture all notions of bias, and thus should not be the only consideration for serving models. While this is a valid risk, this is one that is not specific to prediction sensitivity. Good use of this metric requires users to be cognizant of these strengths and weaknesses. 
We also note that the metric requires defining protected attributes (e.g. gender) and our work carries the limitation that the selected datasets contain binary gender annotations.
Defining protected attributes may not always be possible and when possible, the protected attribute classes may not be comprehensive. 

% TODO : Add limitations and future work.

% The first line of the file must be
% \begin{quote}
% \begin{verbatim}
% \documentclass[11pt]{article}
% \end{verbatim}
% \end{quote}

% To load the style file in the review version:
% \begin{quote}
% \begin{verbatim}
% \usepackage[review]{emnlp2021}
% \end{verbatim}
% \end{quote}
% For the final version, omit the \verb|review| option:
% \begin{quote}
% \begin{verbatim}
% \usepackage{emnlp2021}
% \end{verbatim}
% \end{quote}

% To use Times Roman, put the following in the preamble:
% \begin{quote}
% \begin{verbatim}
% \usepackage{times}
% \end{verbatim}
% \end{quote}
% (Alternatives like txfonts or newtx are also acceptable.)

% Please see the \LaTeX{} source of this document for comments on other packages that may be useful.

% Entries for the entire Anthology, followed by custom entries
\bibliography{anthology,custom,nlp}
\bibliographystyle{acl_natbib}

\newpage
\appendix

\section{Obtaining prediction sensitivity on classifier trained for statistical parity}

Let us consider a classification task on whether to hire a person given the following features: $x_1$ is the person's educational experience in years, $x_2$ is their hair length and $x_3$ is their gender.
We synthetically generate data for individuals in this dataset. 
$x_1$ is drawn uniformly randomly between 0 and 10.
$x_3$ is (again) considered to be binary gender (set 0 for male and 1 for female drawn from a bernoulli distribution) and $x_2$ is drawn from a Gaussian distribution conditioned on $x_3$. $x_2 \sim \mathcal{N}(2,10)$ (Gaussian distribution with a mean 2 and variance 10) if $x_3 = 0$ and $x_2 \sim \mathcal{N}(10,10)$ if $x_3 = 1$.
We sample 10,000 data-points from the above distribution to generate a dataset.
Let us consider two cases with two different classifiers.

\textbf{Case 1: Classifier depends on $x_1, x_2$}
In this case, the modeler only deems $x_3$ to be the protected feature. Let us assume that they build a classifier as shown in equation~\ref{eq:two_clf}.
Lets assume that the modeler assigns a hire decision if ${f} > 0.5$, otherwise not. 

\begin{equation}\label{eq:two_clf}
f = \sigma((x_1 - 5) + (x_2 - 6))
\end{equation}

Given only $x_3$ is considered as the protected feature by the modeler, they will set the vector $\bm v$ to $[0, 0, 1]^T$. Let us assume that the modeler sets $P$ as 
\begin{equation}\label{eq:p_case2}
P = 
\begin{bmatrix} \frac{1}{2} & \frac{1}{2}  \end{bmatrix}
\begin{bmatrix}
\big\lvert \frac{\partial {f_1}}{\partial  x_1} \big\rvert & \big\lvert \frac{\partial {f_1}}{\partial  x_2} \big\rvert & \big\lvert \frac{\partial {f_1}}{\partial  x_3} \big\rvert\\
\big\lvert \frac{\partial {f_2}}{\partial  x_1} \big\rvert & \big\lvert \frac{\partial {f_2}}{\partial  x_2} \big\rvert & \big\lvert \frac{\partial {f_2}}{\partial  x_3} \big\rvert
\end{bmatrix}
\begin{bmatrix} 0 \\ 0  \\ 1  \end{bmatrix}
%\end{align}
\end{equation}

We recommend the modeler computes $\frac{\partial {x_2}}{\partial  x_3}$ and $\frac{\partial {x_1}}{\partial  x_3}$ and if they are non-zero, use the chain rule in equation~\ref{eq:chain_rule} to compute $P$. 

\begin{equation}\label{eq:chain_rule}
\frac{\partial {f_k}([x_1, x_2])}{\partial  x_3} = \frac{\partial {f_k}([x_1, x_2])}{\partial  x_2}  \frac{\partial x_2}{\partial  x_3}
\end{equation}

For the dataset generated above, we compute the partials $\frac{\partial {x_2}}{\partial  x_3}$ and $\frac{\partial {x_1}}{\partial  x_3}$.
Additionally, since $x_3$ is a discrete variable, we approximate partial derivatives using all available right-difference quotients and left-difference quotients, as shown in equation~\ref{eq:diff_quot}. 
In order to compute $\frac{\partial {x_2}}{\partial  x_3}$ at $x_3 = x_3^m$ (where $x_3^m$ denotes the value of $x_3$ for the $m^\text{th}$ data-point), we use the corresponding value of the feature $x_2 = x_2^m$ in the $m^\text{th}$ data-point and all other available pairs $(x_2^n, x_2^n), n \neq m$. 

\begin{equation} \label{eq:diff_quot}
    \frac{\partial {x_2}}{\partial  x_3} \Big|_{x_3 = x_3^m} =  \text{Mean} \Big(\frac{x_2^m - x_2^n}{x_3^m - x_3^n}\Big)
\end{equation}

The mean above is computed over all $n \neq m$. Similarly,
\begin{equation} \label{eq:diff_quot}
    \frac{\partial {x_1}}{\partial  x_3} \Big|_{x_3 = x_3^m} =  \text{Mean} \Big(\frac{x_1^m - x_1^n}{x_3^m - x_3^n}\Big)
\end{equation}

Given the dataset we generated, we compute values for $\frac{\partial {x_1}}{\partial  x_3} \Big|_{x_3 = x_3^m}$ and $\frac{\partial {x_2}}{\partial  x_3} \Big|_{x_3 = x_3^m}$ for an arbitrarily chosen $m$. 
We obtain values of 7.98 and 0.01, respectively. 
Note that we expect the second value to be 0, but due to noise in gradient approximation obtain a non-zero value. 
We re-write equation~\ref{eq:p_case2} as shown below and plug in the values of the partials. We obtain a non-zero value of $P$ in this case. 

\begin{align}\label{eq:example}
    P & = \begin{bmatrix} \frac{1}{2} & \frac{1}{2}  \end{bmatrix}
\begin{bmatrix}
\big\lvert \frac{\partial {f_1}}{\partial  x_1} \big\rvert & \big\lvert \frac{\partial {f_1}}{\partial  x_2} \big\rvert & \big\lvert \frac{\partial {f_1}}{\partial  x_3} \big\rvert\\
\big\lvert \frac{\partial {f_2}}{\partial  x_1} \big\rvert & \big\lvert \frac{\partial {f_2}}{\partial  x_2} \big\rvert & \big\lvert \frac{\partial {f_2}}{\partial  x_3} \big\rvert
\end{bmatrix}
\begin{bmatrix} 0 \\ 0 \\ 1 \end{bmatrix}
 \\ & = 
\begin{bmatrix} \frac{1}{2} & \frac{1}{2}  \end{bmatrix}
\begin{bmatrix}
\big\lvert \frac{\partial {f_1}}{\partial  x_1} \big\rvert & \big\lvert \frac{\partial {f_1}}{\partial  x_2} \big\rvert & \big\lvert \frac{\partial {f_1}}{\partial  x_2}\frac{\partial {x_2}}{\partial  x_3} \big\rvert\\
\big\lvert \frac{\partial {f_2}}{\partial  x_1} \big\rvert & \big\lvert \frac{\partial {f_2}}{\partial  x_2} \big\rvert & \big\lvert \frac{\partial {f_2}}{\partial  x_2}\frac{\partial {x_2}}{\partial  x_3} \big\rvert
\end{bmatrix}
\begin{bmatrix} 0 \\ 0 \\ 1 \end{bmatrix}
\end{align}

\textbf{Case 2: Classifier only depends only on $x_1$}
In this case, the modeler deems both $x_2, x_3$ to be protected features and builds a classifier as depicted below. 

\begin{equation}
f = \sigma (x_1 - 5)
\end{equation}
% \section{AMT experiment setup}
% \label{sec:appendix}

Lets assume that the modeler assigns a hire decision if ${f} > 0.5$, otherwise not. 
Additionally, given $x_2$ and $x_3$ are protected features, $P$ is set to

\begin{equation}\label{eq:p_case1}
P = 
\begin{bmatrix} \frac{1}{2} & \frac{1}{2}  \end{bmatrix}
\begin{bmatrix}
\big\lvert \frac{\partial {f_1}}{\partial  x_1} \big\rvert & \big\lvert \frac{\partial {f_1}}{\partial  x_2} \big\rvert & \big\lvert \frac{\partial {f_1}}{\partial  x_3} \big\rvert\\
\big\lvert \frac{\partial {f_2}}{\partial  x_1} \big\rvert & \big\lvert \frac{\partial {f_2}}{\partial  x_2} \big\rvert & \big\lvert \frac{\partial {f_2}}{\partial  x_3} \big\rvert
\end{bmatrix}
\begin{bmatrix} 0 \\ \frac{1}{2}  \\ \frac{1}{2}  \end{bmatrix}
%\end{align}
\end{equation}

Given that the classifier does not explicitly rely on $x_2$ and $x_3$, we can rewrite equation~\ref{eq:p_case1} as

\begin{equation}\label{eq:p_case1}
P = 
\begin{bmatrix} \frac{1}{2} & \frac{1}{2}  \end{bmatrix}
\begin{bmatrix}
\big\lvert \frac{\partial {f_1}}{\partial  x_1} \big\rvert & \big\lvert \frac{\partial {f_1}}{\partial  x_1}\frac{\partial x_1}{\partial  x_2} \big\rvert & \big\lvert \frac{\partial {f_1}}{\partial  x_1}\frac{\partial x_1}{\partial  x_3} \big\rvert\\
\big\lvert \frac{\partial {f_2}}{\partial  x_1} \big\rvert & \big\lvert \frac{\partial {f_2}}{\partial  x_1}\frac{\partial x_1}{\partial  x_2} \big\rvert & \big\lvert \frac{\partial {f_2}}{\partial  x_1}\frac{\partial x_1}{\partial  x_3} \big\rvert
\end{bmatrix}
\begin{bmatrix} 0 \\ \frac{1}{2}  \\ \frac{1}{2}  \end{bmatrix}
%\end{align}
\end{equation}

We obtain the partial derivatives $\frac{\partial {x_1}}{\partial  x_2} \Big|_{x_2 = x_2^m}$ and $\frac{\partial {x_1}}{\partial  x_3} \Big|_{x_3 = x_3^m}$.
For an arbitrary chosen $x_1^m$, we obtain values of 0.01 and -0.01.
While we expect both these values to be zero given our data construction, they are non-zero due to the gradient approximation.
Barring the noise in gradient computation, $P$ is 0 in this case.

\section{Prediction sensitivity for classifier trained with individual fairness}

We conduct a simulation, where we obtain the proposed metric for increasing values of $L$.
We generate a synthetic dataset with a single feature drawn uniformly randomly between 0 and 10.
The label $y$ of a given datapoint is set to 0 if the feature value is less than 5 or 1 otherwise.
Let us assume we build a linear classifier $f = \theta x$, where $x$ is denotes scalar feature. 
We optimize equation ~\ref{eq:const_opt} and obtain value of $\theta$ that satisfies the constraint and minimizes a chosen $\mathcal L$. 
Let $\mathcal D$ and $d$ be L$_1$ norms and $\mathcal L = (y - f)^2$.
We optimize for the value of $\theta$, and Figure shows the value of accumulated prediction sensitivity with increasing value of $L$ between the range 0 to 0.2.
We observe the metric closely follows value of $L$ till 0.1.
We note that $L$ will equal $\theta$ in this case and the optimal value of $\theta$ in the absence of any constraint is $0.1$.

\begin{figure}[t]
  \centering
   \includegraphics[scale=0.38]{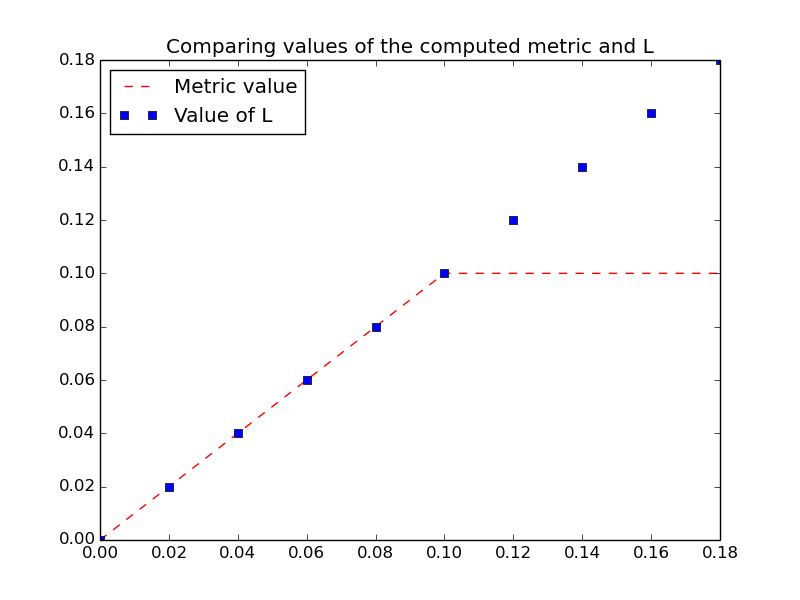}
   \caption{Plot showing the values of the accumulated prediction sensitivity and $L$}
   \label{fig:Mturk}
\end{figure}

\section{Dataset Statistics}

\begin{table}[!ht]
\centering
\begin{tabular}{l c c c}
\toprule
\textbf{Dataset}   & \textbf{Train} & \textbf{Valid} & \textbf{Test}   \\ 
\midrule
\textsc{BiasInBios}     & 107,171 & 71,447 & 91,917 \\
\textsc{JigsawToxi.} & 1,443,900 & 360,974 & 97,320 \\
\bottomrule
\end{tabular}
\caption{Dataset statistics}
\label{table:dataset_stats}
\end{table}

\section{Considerations for using PSM Classifier}
\label{section:appendix_psm_classifier}
Training a PSM classifier requires access to gender labels which might not be available for the dataset used to train the model under evaluation. To overcome this, we evaluate training a PSM classifier on a different dataset and then applying it on the dataset of interest. In Table \ref{tab:table2}, the last two rows record the correlation and mutual information values of a PSM classifier trained on Bias in Bios (tested on Jigsaw) and trained on Jigsaw Toxicity (tested on Bias in Bios), respectively. 
While we beat the CF baseline using the PSM trained on another dataset, comparison to the setting where $\bm v$ is set using gendered words presents a mixed picture.
P3 ($\bm v$ set using PSM trained on Jigsaw Toxicity) has a slightly higher correlation of 0.365 compared to 0.363 in the P5 setting.
However, P3 has a slightly worse MI of 0.091 compared to P5. 
The related experiment for Jigsaw toxicity where $\bm v$ is set using PSM trained on Bias in Bios yields similar mixed observations when compared to P5. 

We also conducted a synthetic experiment wherein we deliberately add bias to the PSM classifier. We reduce the number of `female' datapoints by 50\% leading to about 18\% reduction in the recall for the `female' class (while the `male' class accuracy remains the same). We observe that the metric quality also degrades in this case, leading to a correlation of 0.259 with human judgement, in case of the Bias in Bios data. This correlation is worse than the CF baseline.

Given these results, we observe that using the PSM classifier improves upon other baselines only when it is relatively un-biased in performance across genders and matched to the dataset at hand.
Therefore, we recommend setting $\bm v$ using gendered words if a strong PSM classifier is difficult to obtain.

\begin{table*}[]
\centering
\small
\begin{tabular}{p{7cm}p{1cm}p{1cm}|p{1cm}p{1cm}p{1cm}p{1cm}p{0.9cm}}

\toprule
\multirow{1}{*}{Individual Fairness Metrics} & \multicolumn{2}{c|}{Bias in Bios} & \multicolumn{2}{c}{Jigsaw Toxicity} \\
\midrule
 & Corr. & MI & Corr. & MI  \\ \cline{2-5}
P5 ($\bm v$ set using gendered words and embedding vectors) & \textbf{0.363} & \textbf{0.098} & 0.295 & 0.061  \\ 
P3  ($\bm v$ set using PSM and embedding vectors) & \textbf{0.441} & \textbf{0.105}&  \textbf{0.374} &  \textbf{0.101} \\ 
P3  ($\bm v$ set using PSM(Bias in Bios) and embedding vectors) & & &  \textbf{0.238} &  0.083 \\ 
P3  ($\bm v$ set using PSM(Jigsaw Toxicity) and embedding vectors) & \textbf{0.365} & \textbf{0.091}&   &   \\

\bottomrule

\end{tabular}
\caption{Point bi-serial correlations (Corr.) and Mutual Information (MI) between different individual fairness metrics with human annotations on Bios in Bias and Jigsaw toxicity datasets. Bold numbers are the correlations where we see statistically significant increase over CF baseline. The metric variants are sorted based on the correlation values. We use the bootstrap method to compute statistical significance \cite{koehn2004statistical} at p-value<0.05.}
\label{tab:table2}
\end{table*}

\end{document}